\documentclass[lettersize,journal]{IEEEtran}
\usepackage{amsmath,amsfonts}
\usepackage{algorithmic}
\usepackage{algorithm}
\usepackage{array}
\usepackage[caption=false,font=normalsize,labelfont=sf,textfont=sf]{subfig}
\usepackage{textcomp}
\usepackage{stfloats}
\usepackage{url}
\usepackage{verbatim}
\usepackage{graphicx}
\usepackage{cite}
\usepackage{enumitem}
\usepackage{multirow}
\usepackage{multicol}
\usepackage{amsmath,amssymb,mathrsfs}
\usepackage{booktabs}

\usepackage{booktabs}
\usepackage{makecell}
\usepackage[table]{xcolor}

\usepackage{
amsfonts,
cite,
pifont,
graphicx,
multirow,
booktabs,
amsmath,
amssymb,
hhline,
url,
hyperref}
\usepackage[table,xcdraw]{xcolor}

\usepackage{xcolor}
\definecolor{darkblue}{rgb}{0, 0, 0.6}
\definecolor{darkgreen}{rgb}{0, 0.7, 0}
\definecolor{darkred}{rgb}{0.8, 0, 0}

\let\oldcite\cite
\renewcommand{\cite}[1]{\textcolor{darkblue}{\oldcite{#1}}}

\newcommand{\best}[1]{\textbf{\textcolor{red}{#1}}}

\newcommand*\colorcheck{%
  \expandafter\newcommand\csname greencheck\endcsname{\textcolor{darkgreen}{\ding{52}}}%
}
\newcommand*\colorcross{%
  \expandafter\newcommand\csname redcross\endcsname{\textcolor{darkred}{\ding{56}}}%
}
\colorcheck
\colorcross

\begin{document}

\title{LoCAtion: Long-time Collaborative Attention Framework for High Dynamic Range Video Reconstruction}

\author{Qianyu Zhang, Bolun Zheng*,~\IEEEmembership{Member, IEEE}, Lingyu Zhu, Aiai Huang, \\ 
Zongpeng Li,~\IEEEmembership{Senior Member, IEEE}, 
Shiqi Wang,~\IEEEmembership{Senior Member, IEEE}

\thanks{Qianyu Zhang, Bolun Zheng, Aiai Huang, Zongpeng Li are with the School of Automation, Hangzhou Dianzi University, Hangzhou 310018, China (e-mail: qyzhang@hdu.edu.cn; blzheng@hdu.edu.cn; zongpeng@tsinghua.edu.cn).
\\
Lingyu Zhu and Shiqi Wang are with the Department of Computer Science, City University of Hong Kong (e-mail:lingyzhu-c@my.cityu.edu.hk; shiqwang@cityu.edu.hk)
}
}

\markboth{Journal of \LaTeX\ Class Files,~Vol.~14, No.~8, August~2021}%
{Shell \MakeLowercase{\textit{et al.}}: A Sample Article Using IEEEtran.cls for IEEE Journals}


\maketitle

\begin{abstract}
High dynamic range (HDR) video reconstruction has been widely researched under alternating-exposure inputs, where it is commonly formulated as a frame-centric alignment-and-fusion problem. Although this paradigm has achieved strong performance under controlled conditions, it remains sensitive to large motion, occlusion, saturation, and severe exposure-induced appearance changes. Recently, the dual-stream HDR acquisition setting has emerged to provide a temporally continuous medium-exposure stream as a reliable structural anchor. Yet, effectively incorporating complementary low-/high-exposure cues from the alternating stream without re-introducing fragile alignment errors remains challenging.
In this paper, we propose LoCAtion, a Long-time Collaborative Attention framework tailored to dual-stream HDR video reconstruction. Our key idea is to convert the hardware-level continuity of the medium-exposure stream into an algorithmic reconstruction prior. Specifically, LoCAtion formulates HDR video reconstruction as backbone-guided cross-exposure information compensation: the continuous medium-exposure stream preserves coherent scene structure, while low-/high-exposure anchors provide complementary dynamic-range information. To this end, Collaborative Feature Attention (CFA) stage routes reliable cross-exposure cues into the medium-exposure backbone without explicit optical-flow-based warping, while Global Sequence Consistency (GSC) stage propagates bidirectional and long-range context to refine the reconstructed sequence. Extensive experiments on synthetic and real-world benchmarks demonstrate that LoCAtion achieves leading reconstruction quality and temporal stability, especially on fast-motion and complex scenes, while maintaining a competitive balance between reconstruction accuracy and computational efficiency. Additional results are available at: \url{https://zqqqyu.github.io/Location/}.
\end{abstract}

\begin{IEEEkeywords}
High Dynamic Range Video Reconstruction, Temporally Coherent, Motion-Aware.
\end{IEEEkeywords}

\section{Introduction}

\IEEEPARstart{H}{igh} dynamic range video acquisition seeks to faithfully capture the luminance of real-world scenes from deep shadows to intense highlights \cite{debevec2008recovering, yao2023bidirectional}. While specialized hardware sensors \cite{jiang2021hdr, han2020neuromorphic} directly expand the dynamic range, they often incur prohibitive costs and lack the flexibility required for widespread deployment. Therefore, computational approaches have become highly practical alternatives to overcome these hardware limitations.

\begin{figure}[!th] 
\centering 
\includegraphics[width=0.5\textwidth]{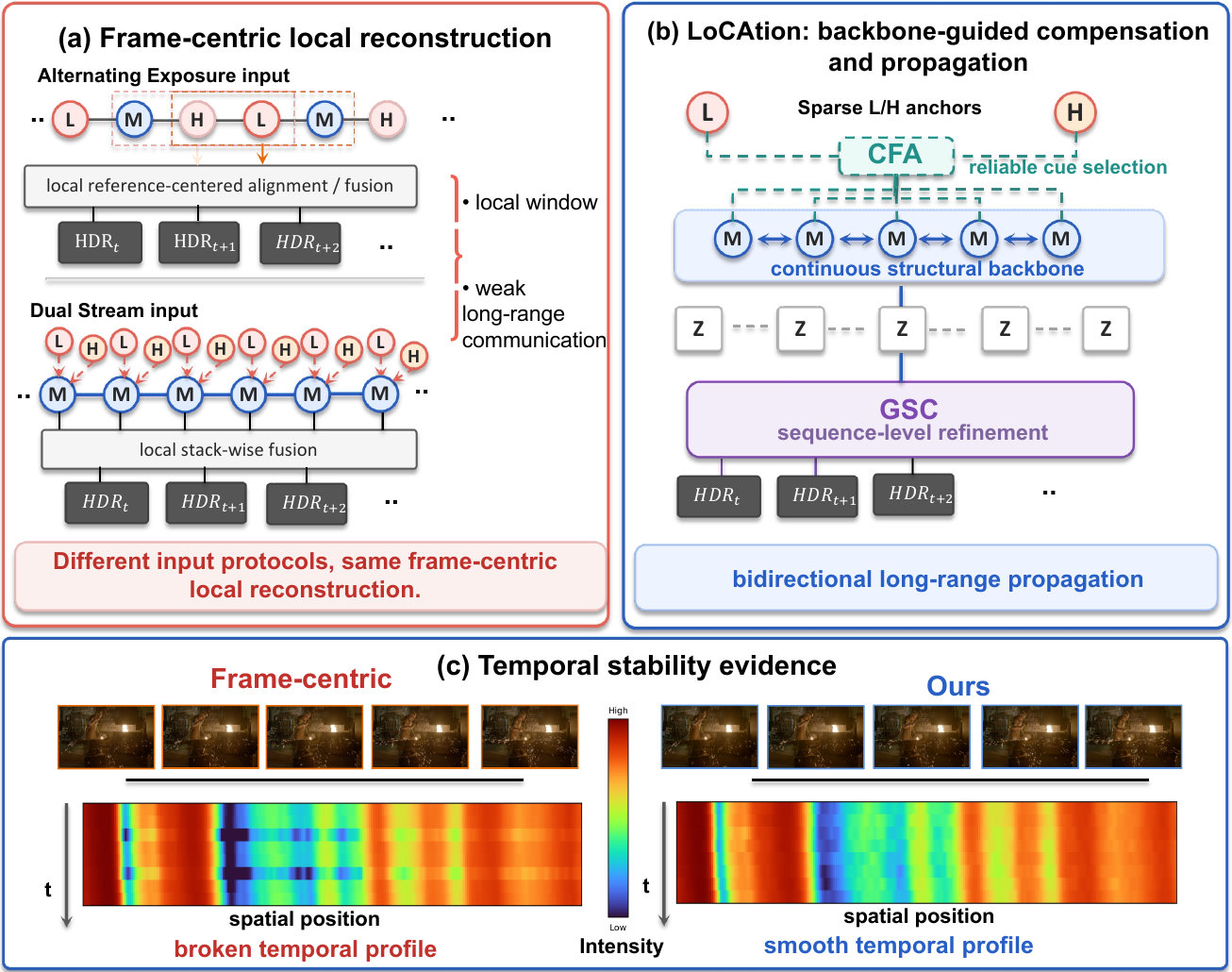} 
\caption{\textbf{Motivation of LoCAtion.}
(a) Existing HDR video reconstruction pipelines (alternating-exposure or dual-stream) are largely formulated as frame-centric local reconstruction, recovering each HDR frame from a short reference-centered window. 
(b) LoCAtion reformulates dual-stream HDR reconstruction as backbone-guided compensation and propagation. 
(c) Temporal profile visualization: intensity values sampled along a fixed scanline are stacked across consecutive frames, with horizontal axis as spatial position and vertical axis as time. Broken/stripe-like profiles indicate frame-to-frame intensity inconsistency or temporal instability, while smooth and coherent profiles reflect improved temporal stability.}
\label{videomef} 
\label{fig:motivation}
\end{figure}

Most existing HDR video methods adopt an alternating-exposure setting and reconstruct each HDR frame from a short temporal neighborhood through local alignment and fusion \cite{kang2003high, kalantari2019deep, yin2021two, chen2021hdr, cui2024exposure, xu2024hdrflow, huang2026ug}. As illustrated in Fig.~\ref{fig:motivation}(a), these methods are typically frame-centric: neighboring LDR frames are aligned or implicitly associated with a reference frame, and then fused to recover missing dynamic-range information. This local reconstruction paradigm has achieved strong performance, but it remains challenging in complex dynamic scenes. On the one hand, cross-exposure correspondence is difficult to establish reliably under large motion, occlusion, saturation, and severe exposure-induced appearance changes. Explicit optical-flow warping may fail in these cases \cite{kalantari2019deep}, while implicit alignment strategies, such as deformable convolution \cite{chen2021hdr, tan2021deep} and cross-frame attention \cite{chung2023lan}, reduce the dependence on explicit flow estimation but involve computationally demanding offset prediction, dynamic sampling, or cross-frame computation. On the other hand, since each output frame is reconstructed within a local window, temporal information is only weakly propagated across the sequence, making it difficult to correct local ambiguities using broader temporal context. As a result, frame-centric reconstruction may suffer from unstable temporal appearance in challenging dynamic regions. The temporal profiles in Fig.~\ref{fig:motivation}(c) provide a visual example of such frame-to-frame inconsistency.

Recently, dual-stream HDR acquisition has introduced a continuous medium-exposure stream together with an alternating low-/high-exposure stream \cite{zhang2025capturing}. Compared with conventional alternating-exposure capture, this setting provides a different source of temporal reliability. 
However, directly adapting conventional fusion pipelines to this input does not fully realize this potential. As shown in the lower part of Fig.~\ref{fig:motivation}(a), such a method still follows a frame-centric reconstruction paradigm, where each target frame is mainly recovered through local cross-exposure fusion around the current time step. Although the medium-exposure frames are continuously available, conventional local fusion mainly uses them as per-frame or local-window references, rather than exploiting their continuity as a persistent structural cue across the reconstruction segment. As a result, the main advantage of dual-stream acquisition, i.e., the persistent medium-exposure structure across time, remains underexploited.

To address this issue, we propose LoCAtion, a Long-time Collaborative Attention framework for dual-stream HDR video reconstruction. Instead of using dual-stream inputs only for local exposure fusion, LoCAtion exploits the continuous medium-exposure stream as a temporal structural backbone. This changes the reconstruction focus from reference-centered alignment and fusion to backbone-guided compensation and sequence-level refinement. The medium-exposure stream maintains coherent scene structure across time, while the sparse low-/high-exposure frames provide complementary dynamic-range cues for regions that are saturated or under-exposed.
Based on this formulation, LoCAtion is designed to answer two practical questions: how to inject useful exposure information without relying on fragile cross-exposure warping, and how to propagate the reconstructed information along the continuous backbone. For the first question, we introduce Collaborative Feature Attention (CFA) stage, which performs reliability-guided cross-exposure cue routing and selectively injects complementary low-/high-exposure features into the medium-exposure backbone. For the second question, we introduce Global Sequence Consistency (GSC) stage, which refines local HDR estimates by aggregating reconstruction context from both temporal directions and modeling dependencies within the reconstruction segment. In this way, LoCAtion turns the continuity provided by dual-stream acquisition into an explicit reconstruction prior, leading to more stable HDR recovery in dynamic scenes.

We summarize our main contributions as follows:
\begin{itemize}
\item 
We convert the hardware-level continuity of dual-stream acquisition into an algorithmic prior, where the continuous medium-exposure stream serves as a structural backbone and sparse low-/high-exposure anchors provide complementary dynamic-range cues, shifting the reconstruction from frame-centric local fusion toward continuity-aware backbone-guided compensation.

\item 
We develop LoCAtion to exploit this prior through reliability-guided cross-exposure cue selection and sequence-level temporal propagation. This design avoids explicit optical-flow-based warping of low-/high-exposure anchors and refines local HDR estimates along the continuous medium-exposure backbone.

\item 
Experiments on synthetic benchmarks and real-captured HDR videos demonstrate that LoCAtion improves reconstruction quality and temporal stability, especially in fast-motion and complex scenes, while maintaining a favorable accuracy-efficiency trade-off.

\end{itemize}

\section{Related Work}
\label{sec: related work}

\noindent \textbf{Multi-exposure HDR Image Reconstruction}. 
HDR image reconstruction aims to recover high dynamic range radiance from multiple LDR images captured with different exposures \cite{nam2024deep}. Early methods mainly relied on hand-crafted alignment or motion rejection strategies to suppress ghosting artifacts. Alignment-based approaches estimated motion using optical flow \cite{bogoni2000extending}, patch-based optimization \cite{sen2012robust}, or luminance optimization \cite{hu2013hdr}. Rejection-based methods detected unreliable moving regions according to intensity differences \cite{grosch2006fast}, weighted variance \cite{jacobs2008automatic}, or graph-cut optimization \cite{zhang2011gradient}. Although effective in simple cases, these methods often struggle with large motion, occlusions, saturation, and complex dynamic scenes.

Deep learning has significantly advanced multi-exposure HDR image reconstruction \cite{hu2022hdr, jo2021deep}. Kalantari \textit{et al.} \cite{kalantari1} first introduced a CNN-based pipeline that uses optical flow for alignment and learns to fuse multi-exposure inputs. Subsequent works improved robustness by learning alignment and fusion jointly, including flow-based methods \cite{peng2018deep, prabhakar2019fast, kong2024safnet}, attention-based methods such as AHDNet \cite{yan2019attention} and ADNet \cite{liu2021adnet}, and Transformer-based methods such as HDR-Trans \cite{liu2022ghost} and HyHDRNet \cite{yan2023unified}. These methods demonstrate the importance of adaptive correspondence modeling under exposure variation. However, HDR image reconstruction is inherently frame-level and does not address long-range temporal coherence, which becomes critical in HDR video reconstruction.

\noindent \textbf{Alignment-centric HDR Video Reconstruction}. 
Computational HDR video reconstruction commonly recovers HDR sequences from LDR frames captured with alternating exposures, avoiding the need for expensive dedicated HDR hardware \cite{nayar2000camera, choi2017reconstructing}. Kang \textit{et al.} \cite{kang2003high} introduced the alternating-exposure paradigm, where neighboring frames are aligned and merged to reconstruct HDR video frames. Traditional approaches improved motion handling through block-based motion estimation \cite{mangiat2010high}, HDR filtering \cite{mangiat2011spatially}, adaptive weighting \cite{gryaditskaya2015motion}, or statistical fusion without exact alignment \cite{li2016maximum}. Nevertheless, these methods are often sensitive to inaccurate motion estimation and may produce visible artifacts in complex scenes.

Recent deep HDR video methods largely inherit the frame-centric alignment-and-fusion formulation. Kalantari \textit{et al.} \cite{kalantari2019deep} aligned alternating-exposure frames using optical flow and fused them with an end-to-end network. Chen \textit{et al.} \cite{chen2021hdr} adopted deformable convolution in a coarse-to-fine framework to handle complex motions. LAN-HDR \cite{chung2023lan} used luminance-aware attention to associate adjacent frames with the reference, while HDRFlow \cite{xu2024hdrflow} designed an efficient flow network for large-motion alignment. Cui \textit{et al.} \cite{cui2024exposure} further introduced exposure completion and feature interpolation to improve alignment across exposure differences. Although these methods differ in their alignment mechanisms, most of them reconstruct each HDR frame within a local reference-centered window. As a result, their performance still depends on reliable frame-wise correspondences, and temporal information is only weakly propagated across long sequences.


\noindent \textbf{Dual-stream HDR Reconstruction and Temporal Propagation}. 
Recent dual-stream HDR acquisition provides a continuous medium-exposure stream and an alternating-exposure stream \cite{zhang2025capturing}, offering a stable structural backbone and complementary dynamic-range cues, respectively. However, conventional frame-centric alignment-and-fusion pipelines do not fully exploit this temporal continuity. Although recurrent or bidirectional propagation has been widely studied in general video restoration \cite{chan2022basicvsr++}, HDR video reconstruction additionally requires handling cross-exposure appearance gaps, saturation, and exposure-dependent information loss. Our method therefore propagates reconstruction context along the continuous medium-exposure backbone and couples it with warping-free cross-exposure cue routing, enabling local dynamic-range recovery and sequence-level temporal refinement in the dual-stream HDR framework.

\begin{figure*}[t] 
\centering 
\includegraphics[width=0.95\textwidth]{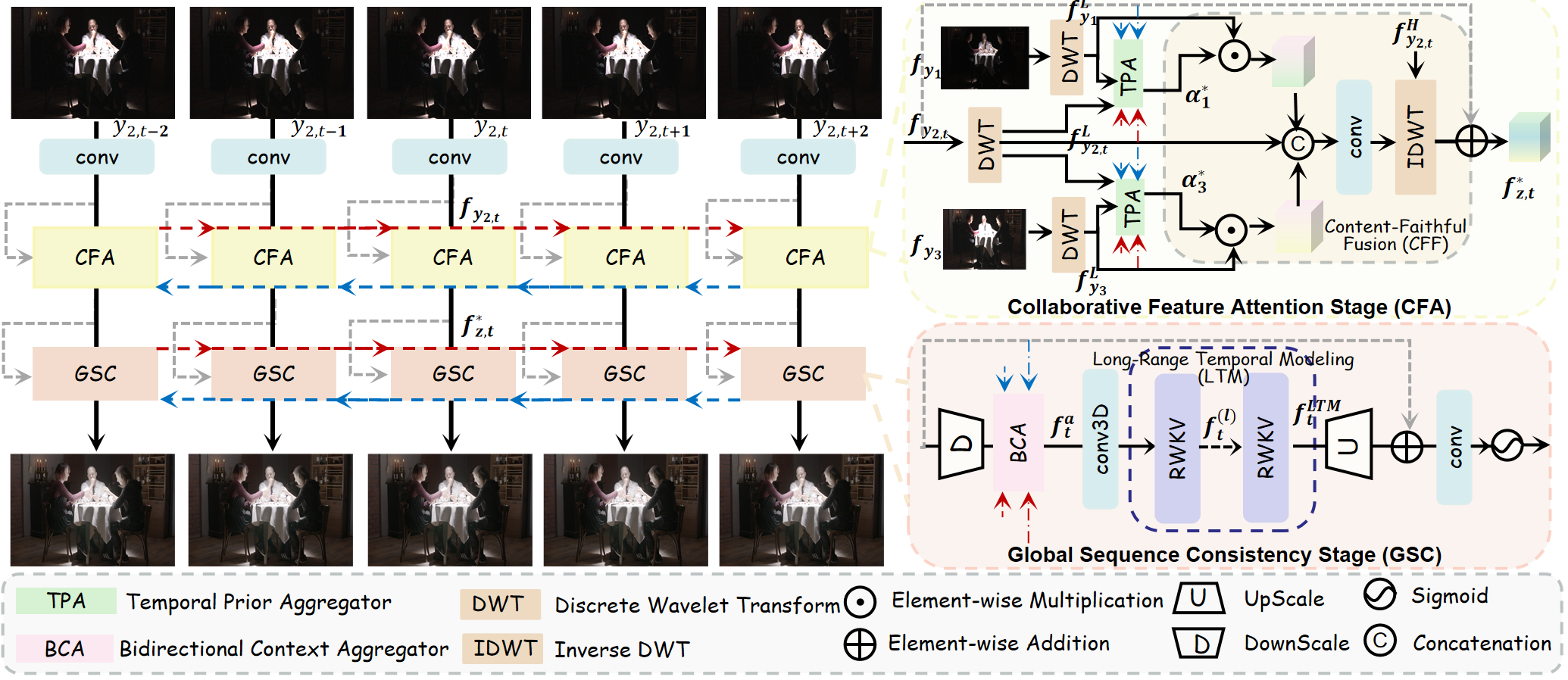} 
\caption{Overall architecture of LoCAtion for dual-stream HDR video reconstruction.
Given a continuous medium-exposure sequence and sparse low-/high-exposure anchors, LoCAtion organizes reconstruction around the medium-exposure backbone.
In the Collaborative Feature Attention (CFA) stage, Temporal Prior Aggregator (TPA) estimates reliability guidance for the low-/high-exposure anchors, and Content-Faithful Fusion (CFF) selectively injects reliable cross-exposure cues in the low-frequency wavelet domain while preserving the structural content of the medium-exposure stream.
The intermediate HDR estimates are further refined by Global Sequence Consistency (GSC), where Bidirectional Context Aggregation (BCA) propagates reconstruction context from both temporal directions and Long-Range Temporal Modeling (LRTM) models temporal dependencies within the selected reconstruction segment.
Red and blue dashed arrows denote forward and backward temporal propagation, respectively.} 
\label{framework} 
\end{figure*}

\section{Methodology}
\label{sec:Methodology}

\subsection{Motivation and Problem Formulation}
\label{sec:Motivation}
Dynamic HDR video reconstruction is challenging because the network must recover missing dynamic-range information while handling inter-frame motion and cross-exposure appearance variations. Conventional frame-centric methods usually couple motion estimation with exposure fusion, making the reconstruction sensitive to inaccurate correspondence estimation under large motion, occlusion, and saturation.

We build LoCAtion upon the dual-stream capture setup of \cite{zhang2025capturing}. For a temporal segment, the input consists of a continuous medium-exposure stream
$\mathbf{Y}_2=\{y_{2,t}\}_{t=1}^{T}$ and a pair of low-/high-exposure anchors $\{y_1,y_3\}$. The medium-exposure stream provides a temporally continuous structural backbone, while the low- and high-exposure anchors provide complementary dynamic-range cues for saturated or under-exposed regions. Instead of explicitly warping the exposure anchors to each medium-exposure frame, we formulate the task as backbone-guided cross-exposure compensation followed by sequence-level temporal refinement.

As illustrated in Fig.~\ref{framework}, LoCAtion contains two stages. First, the Collaborative Feature Attention stage performs local dynamic-range recovery by routing reliable low-/high-exposure cues into the medium-exposure backbone. Second, the Global Sequence Consistency stage propagates reconstruction context along the sequence to refine local HDR estimates with longer-range temporal evidence.

\subsection{Stage 1: Collaborative Feature Attention (CFA)}
\label{sec:FusionStep}

The goal of this stage is to recover local dynamic range by injecting reliable cross-exposure information into the continuous medium-exposure backbone. Existing HDR video methods usually establish correspondences between differently exposed frames through optical flow, deformable convolution, or attention-based matching. However, in dynamic scenes, intensity differences may be caused by both object motion and exposure changes, making frame-wise correspondence estimation unreliable and prone to ghosting artifacts.

In the dual-stream setting, the continuous medium-exposure sequence $\mathbf{Y}_2$ provides a more stable structural reference. We therefore do not use the low- and high-exposure anchors $\{y_1,y_3\}$ as motion references. Instead, they are treated as complementary dynamic-range sources for saturated or under-exposed regions. Based on this design, Stage 1 performs reliability-guided feature routing rather than explicit optical-flow-based warping. Specifically, it estimates where the anchor information is reliable for each medium-exposure frame and selectively injects valid dynamic-range cues into the backbone representation.

Since high-frequency details are sensitive to motion, saturation, and exposure-induced appearance changes, directly transferring them across exposures may introduce unstable responses. Therefore, we estimate cross-exposure reliability on the low-frequency components and preserve medium-exposure high-frequency information for reconstruction. Given $y_i\in\mathbb{R}^{3\times H\times W}$ for $i\in\{1,3\}$ and $y_{2,t}\in\mathbb{R}^{3\times H\times W}$, we first extract feature maps and apply DWT to obtain the low-frequency components:
\begin{equation}
f_i^{L}=\mathrm{LL}(\mathcal{W}(\mathcal{E}(y_i))), \quad
f_{2,t}^{L}=\mathrm{LL}(\mathcal{W}(\mathcal{E}(y_{2,t}))),
\end{equation}
where $\mathcal{E}(\cdot)$ is the feature encoder, $\mathcal{W}(\cdot)$ denotes DWT, and $f_i^{L}, f_{2,t}^{L}\in\mathbb{R}^{C\times \frac{H}{2}\times \frac{W}{2}}$. 

\noindent \textbf{Temporal Prior Aggregator (TPA).} 
Estimating anchor reliability from a single anchor-frame pair is ambiguous, since their differences may be caused by either exposure variation or motion mismatch. To obtain temporally informed fusion guidance, the TPA predicts time-dependent reliability maps by aggregating context along the continuous medium-exposure backbone.

For each exposure anchor $i\in\{1,3\}$, TPA takes the anchor feature $f_i^{L}$ and the backbone feature sequence $\{f_{2,t}^{L}\}_{t=1}^{T}$ as input. It recurrently aggregates temporal context in both forward and backward directions:
\begin{equation}
\label{eq:tpa_prop}
\begin{aligned}
\overrightarrow{h}_{i,t} &=
\mathcal{G}_{i}^{\rightarrow}
\left(
[\overrightarrow{h}_{i,t-1}, f_i^{L}, f_{2,t}^{L}]
\right), \\
\overleftarrow{h}_{i,t} &=
\mathcal{G}_{i}^{\leftarrow}
\left(
[\overleftarrow{h}_{i,t+1}, f_i^{L}, f_{2,t}^{L}]
\right),
\end{aligned}
\end{equation}
where $\mathcal{G}_{i}^{\rightarrow}$ and $\mathcal{G}_{i}^{\leftarrow}$ are convolutional aggregation blocks, $[\cdot]$ denotes channel-wise concatenation, and the hidden states are initialized as
$\overrightarrow{h}_{i,0}=\mathbf{0}$ and $\overleftarrow{h}_{i,T+1}=\mathbf{0}$.

The reliability map for the $i$-th exposure anchor at time $t$ is then predicted by
\begin{equation}
\label{eq:tpa_alpha}
\alpha_{i,t}
=
\sigma
\left(
\mathcal{P}_{i}
\left(
[\overrightarrow{h}_{i,t}, \overleftarrow{h}_{i,t}, f_i^{L}, f_{2,t}^{L}]
\right)
\right),
\end{equation}
where $\mathcal{P}_{i}$ is a projection layer and $\sigma(\cdot)$ denotes the sigmoid activation. The resulting map
$\alpha_{i,t}\in[0,1]^{C\times \frac{H}{2}\times \frac{W}{2}}$
indicates the feature-level reliability of the anchor $y_i$ for compensating the $t$-th medium-exposure frame.

\noindent \textbf{Content-Faithful Fusion (CFF).} 
Given the reliability maps $\boldsymbol{\alpha} = \{\boldsymbol{\alpha}_{1,t}, \boldsymbol{\alpha}_{3,t}\}$, CFF uses them as soft gates to modulate the contribution of the low- and high-exposure anchors. Reliable anchor features are injected into the medium-exposure backbone for dynamic-range compensation, while regions with potential motion mismatch are suppressed. The fused low-frequency feature is then combined with the high-frequency component of the medium-exposure backbone through inverse DWT to obtain the intermediate estimate $z_t$.

\subsection{Stage 2: Global Sequence Consistency (GSC)}
\label{sec:TemporalStep}

The CFA stage selects and injects complementary low-/high-exposure cues to improve local dynamic-range recovery for each medium-exposure frame. However, residual inconsistencies may still remain among the intermediate estimates, especially in regions with uncertain cross-exposure cues. We therefore introduce GSC to refine the intermediate sequence by propagating reconstruction context along the continuous medium-exposure backbone.

Given the intermediate sequence $\mathbf{Z}=\{z_t\}_{t=1}^{T}$, we predict a spatio-temporal residual $\mathbf{\Delta}=\{\Delta_t\}_{t=1}^{T}$:
\begin{equation}
\label{eq:gsc_residual}
x_t = z_t + \Delta_t,
\quad
\Delta_t = \Psi_{\mathrm{GSC}}(\mathbf{Z})_t,
\end{equation}
where $x_t$ is the final HDR reconstruction. This residual design allows GSC to refine the CFA output through sequence-level correction.

We first encode the intermediate HDR sequence into low-resolution features:
\begin{equation}
f_{t}^{(0)}=\mathcal{E}_{2}(z_t),
\quad t=1,\dots,T.
\end{equation}
Then, two complementary modules are used to aggregate the temporal context.

\noindent \textbf{Bidirectional Context Aggregation (BCA).} 
BCA propagates contextual information in both temporal directions. For the $k$-th pass, the forward and backward hidden features are updated as
\begin{equation}
\label{eq:bca}
\begin{aligned}
\overrightarrow{h}_{t}^{(k)}
&=
\mathcal{B}_{\rightarrow}^{(k)}
\left(
[\overrightarrow{h}_{t-1}^{(k)}, q_{t}^{\rightarrow(k)}]
\right), \\
\overleftarrow{h}_{t}^{(k)}
&=
\mathcal{B}_{\leftarrow}^{(k)}
\left(
[\overleftarrow{h}_{t+1}^{(k)}, q_{t}^{\leftarrow(k)}]
\right),
\end{aligned}
\end{equation}
where $\mathcal{B}_{\rightarrow}^{(k)}$ and $\mathcal{B}_{\leftarrow}^{(k)}$ are residual propagation blocks. For the first pass, we use
$q_{t}^{\rightarrow(1)}=q_{t}^{\leftarrow(1)}=f_t^{(0)}.$
For the second pass, the two directions exchange information:
$q_{t}^{\rightarrow(2)}=[f_t^{(0)}, \overleftarrow{h}_{t}^{(1)}],\quad
q_{t}^{\leftarrow(2)}=[f_t^{(0)}, \overrightarrow{h}_{t}^{(1)}].$
The aggregated feature is obtained as
\begin{equation}
\label{eq:bca_fuse}
f_{t}^{a}
=
\mathcal{P}_{a}
\left(
[f_t^{(0)},
\overrightarrow{h}_{t}^{(1)},
\overleftarrow{h}_{t}^{(1)},
\overrightarrow{h}_{t}^{(2)},
\overleftarrow{h}_{t}^{(2)}]
\right),
\end{equation}
where $\mathcal{P}_{a}$ compresses the concatenated features into a compact representation.

\noindent \textbf{Long-Range Temporal Modeling (LRTM).} Although BCA propagates information recurrently, long-range dependency modeling remains important for improving sequence-level consistency. 
We therefore introduce LRTM using $K$ stacked RWKV blocks \cite{ duan2024vision}. 
Let $r_t^{(0)}=f_t^{a}$. The $l$-th RWKV block processes the whole feature sequence as
\begin{equation}
\label{eq:rwkv}
\{r_t^{(l)}\}_{t=1}^{T}
=
\mathrm{RWKV}^{(l)}
\left(
\{r_t^{(l-1)}\}_{t=1}^{T}
\right),
\quad l=1,\dots,K.
\end{equation}
The final long-range representation is $f_t^{\mathrm{LRTM}}=r_t^{(K)}$. We decode it into the residual correction and obtain the final HDR frame by Eq.~\eqref{eq:gsc_residual}. Through bidirectional propagation and long-range temporal modeling, this stage refines local HDR estimates with sequence-level context and improves temporal coherence. The temporal range of LRTM is controlled by the segment length $T$: increasing $T$ enables longer-range dependency modeling, while also increasing memory consumption. 
In the main experiments, we use $T=5$ as a practical trade-off, and analyze larger values of $T$ in Sec.~\ref{Ablation study}. 

\newcommand{\second}[1]{\underline{#1}}

\begin{table*}[t]
\centering
\caption{
Quantitative comparison on the 16-scene HDR video benchmark.
The benchmark is divided into two subsets: six slow-motion/simple scenes
and ten dynamic/complex scenes.
AE and DS denote alternating-exposure inputs and dual-stream inputs, respectively.
Best and second-best results are highlighted in \best{bold} and \second{underline}.
}
\scriptsize
\setlength{\tabcolsep}{2.6pt}
\renewcommand{\arraystretch}{1.12}
\resizebox{\textwidth}{!}{
\begin{tabular}{ll l cc cccccc cccccc}
\toprule
\multirow{2}{*}{\textbf{Input}}
&
\multirow{2}{*}{\textbf{Framework}}
&
\multirow{2}{*}{\textbf{Method}}
&
\multirow{2}{*}{\makecell[c]{\textbf{Params}\\\textbf{(M)} $\downarrow$}}
&
\multirow{2}{*}{\makecell[c]{\textbf{Time}\\\textbf{(ms)} $\downarrow$}}
&
\multicolumn{6}{c}{\textbf{Slow/Simple Scenes (6)}}
&
\multicolumn{6}{c}{\textbf{Dynamic/Complex Scenes (10)}}
\\
\cmidrule(lr){6-11}
\cmidrule(lr){12-17}
&
&
&
&
&
\makecell[c]{PSNR$_\mu$\\$\uparrow$}
&
\makecell[c]{SSIM$_\mu$\\$\uparrow$}
&
\makecell[c]{$t$-PSNR\\$\uparrow$}
&
\makecell[c]{$t$-SSIM\\$\uparrow$}
&
\makecell[c]{FovVideoVDP\\$\uparrow$}
&
\makecell[c]{STD\\$\downarrow$}
&
\makecell[c]{PSNR$_\mu$\\$\uparrow$}
&
\makecell[c]{SSIM$_\mu$\\$\uparrow$}
&
\makecell[c]{$t$-PSNR\\$\uparrow$}
&
\makecell[c]{$t$-SSIM\\$\uparrow$}
&
\makecell[c]{FovVideoVDP\\$\uparrow$}
&
\makecell[c]{STD\\$\downarrow$}
\\
\midrule

\multirow{3}{*}{\makecell[c]{AE}}
&
\multirow{3}{*}{\makecell[c]{Frame-\\centric}}
&
DeepHDRVideo(ICCV'21) \cite{chen2021hdr}
&
2.76
&
167.60
&
36.78
&
0.8674
&
39.26
&
0.8814
&
9.7251
&
4.30
&
38.71
&
0.9516
&
41.22
&
0.9405
&
9.0058
&
3.95
\\

&
&
LAN-HDR(ICCV'23) \cite{chung2023lan}
&
7.39
&
230.70
&
36.82
&
0.8617
&
39.44
&
0.8827
&9.7265

&
4.23
&
38.64
&
0.9539
&
41.46
&
0.9475
&
9.0568
&
3.52
\\

&
&
HDRFlow(CVPR'24) \cite{xu2024hdrflow}
&
5.33
&
17.60
&
\best{36.93}
&
\second{0.8696}
&
39.39
&
0.8869
&
\second{9.7611}
&
3.90
&
38.52
&
0.9514
&
40.37
&
0.9499
&
8.9856
&
4.29
\\

\midrule

\multirow{7}{*}{\makecell[c]{DS}}
&
\multirow{6}{*}{\makecell[c]{Frame-\\centric}}
&
AHDNet(CVPR'19) \cite{yan2019attention}
&
1.44
&
158.90
&
36.45
&
0.8663
&
40.27
&
0.8852
&
9.6958
&
\second{0.39}
&
39.32
&
\second{0.9630}
&
43.35
&
0.9527
&
9.2846
&
1.68
\\

&
&
DomainPlus(MM'22) \cite{zheng2022domainplus}
&
24.78
&
116.90
&
36.54
&
0.8588
&
40.34
&
0.8833
&
9.7260
&
0.42
&
39.58
&
0.9608
&
43.49
&
0.9510
&
9.2416
&
1.71
\\

&
&
HDRTransformer(ECCV'22) \cite{liu2022ghost}
&
1.46
&
1531.30
&
36.63
&
0.8692
&
40.43
&
\best{0.8879}
&
9.7359
&
0.44
&
39.79
&
0.9628
&
43.73
&
0.9535
&
9.4402
&
1.69
\\

&
&
SCTNet(ICCV'23) \cite{tel2023alignment}
&
0.96
&
2553.90
&
36.70
&
0.8628
&
40.43
&
0.8861
&
9.7316
&
0.40
&
39.86
&
0.9613
&
43.71
&
0.9523
&
9.4268
&
1.67
\\

&
&
AFUNet(ICCV'25) \cite{li2025afunet}
&
1.14
&
2341.50
&
36.59
&
0.8679
&
40.27
&
0.8845
&
9.7151
&
0.43
&
39.71
&
0.9622
&
43.55
&
0.9521
&
9.3708
&
1.69
\\

&
&
EAFNet(arxiv'26) \cite{zhang2025capturing}
&
6.71
&
463.90
&
36.88
&
\best{0.8697}
&
\second{40.47}
&
\second{0.8875}
&
9.7425
&
0.42
&
\second{39.91}
&
0.9627
&
\second{43.82}
&
\second{0.9536}
&
\second{9.4526}
&
\second{1.65}
\\

\cmidrule(lr){2-17}

\rowcolor{gray!10}
&
\makecell[c]{Propagation}
&
\textbf{Ours}
&
4.63
&
82.16
&
\second{36.90}
&
0.8688
&
\best{40.55}
&
\best{0.8879}
& \best{9.7729}

&
\best{0.37}
&
\best{40.41}
&
\best{0.9635}
&
\best{44.09}
&
\best{0.9540}
&
\best{9.5365}
&
\best{1.48}
\\

\bottomrule
\end{tabular}
}
\vspace{1mm}
\label{tab:main_syn}
\end{table*}

\begin{figure}[!th] 
\centering 
\includegraphics[width=0.5\textwidth]{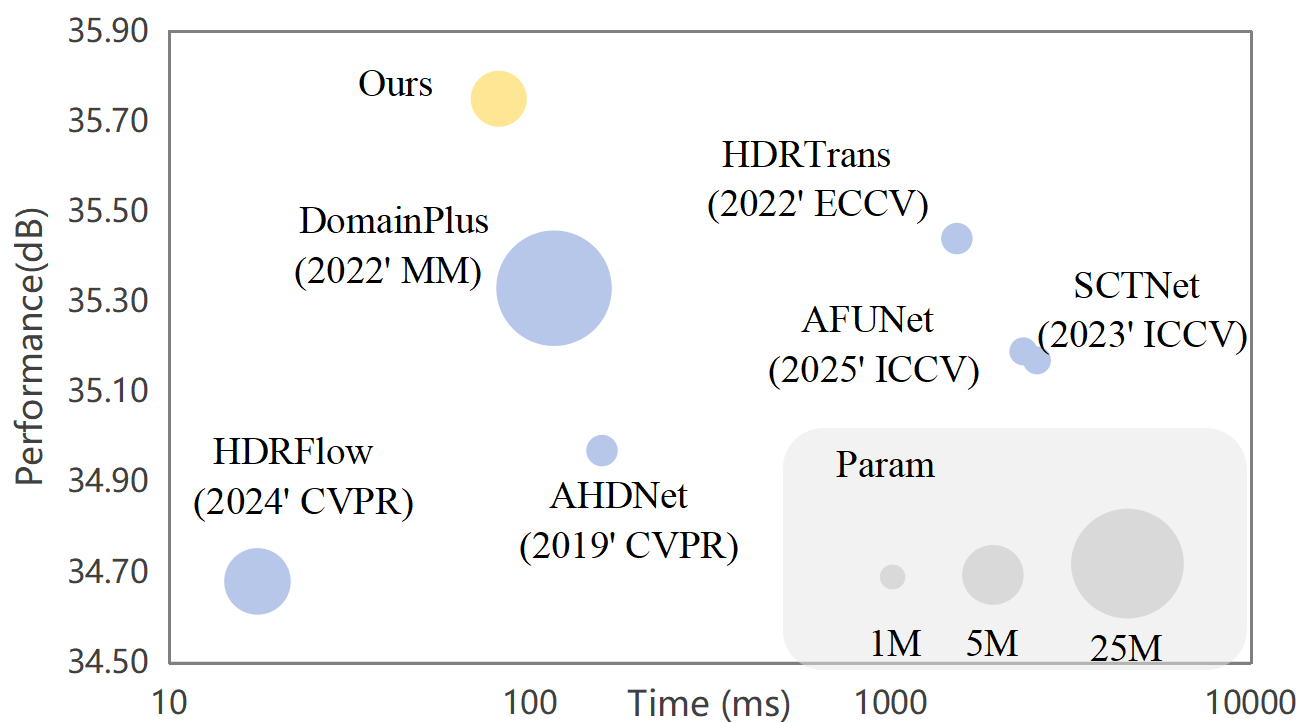} 
\caption{
\textbf{Speed-performance trade-off comparison} with recent state-of-the-art HDR methods on the Cinematic Video dataset. 
LoCAtion attains high reconstruction quality with moderate model size and substantially lower latency than heavy Transformer-based or frame-centric dual-stream baselines.} 
\label{fig:speed_tradeoff} 
\end{figure}

\begin{figure*}[t] 
\centering 
\includegraphics[width=1\textwidth]{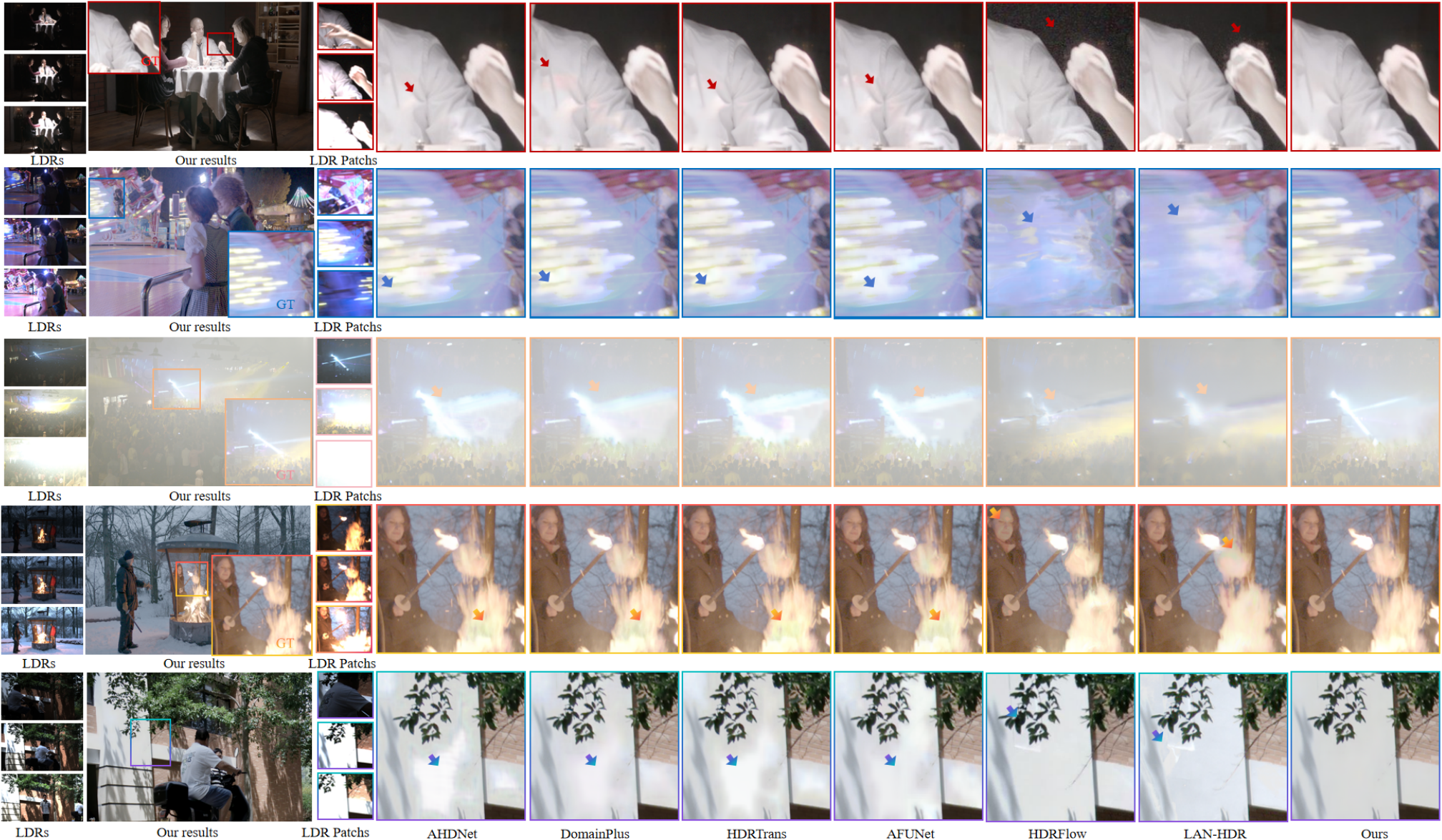} 
\caption{Qualitative comparisons on representative synthetic and real HDR video scenes. 
Each row shows a challenging case involving saturated highlights, complex lighting conditions, or fast motion. 
Image-based methods tend to leave residual ghosts or blurred textures due to insufficient temporal reasoning, while video-based alignment methods may introduce noise or motion-induced artifacts. 
LoCAtion better suppresses ghosting and noise while preserving fine structures in both highlight and shadow regions.} 
\label{figexp1} 
\vspace{-3mm}
\end{figure*}

\subsection{Optimization Objective}

Let $\tau(\cdot)$ define a logarithmic tone-mapping function $\tau(x)=\frac{\log(1+\kappa x)}{\log(1+\kappa)}$ with $\kappa=5000$. We supervise the network using three tailored loss functions. 

We first supervise the final HDR prediction in the tone-mapped domain:
\begin{equation}
\mathcal{L}_{\mathrm{s}} = \frac{1}{T}\sum_{t=1}^{T} \left\| \tau(\widehat{x}_t) - \tau(x_t) \right\|_{1}
\label{eq:loss_rec}
\end{equation}

Second, to explicitly suppress flicker while preserving natural scene motion, we penalize deviations in inter-frame gradients:
\begin{equation}
\mathcal{L}_{\mathrm{t}} = \frac{1}{T-1}\sum_{t=1}^{T-1} \left\| \big(\tau(\widehat{x}_{t+1})-\tau(\widehat{x}_{t})\big) - \big(\tau(x_{t+1})-\tau(x_{t})\big) \right\|_{1}
\end{equation}

To make the Stage-1 reconstruction less sensitive to the specific choice of exposure anchors, we introduce an anchor consistency loss:
\begin{equation}
\mathcal{L}_{\mathrm{anc}} =
\frac{1}{T}\sum_{t=1}^{T}
\left\|
\tau(z_t^{(a)})-\tau(z_t^{(b)})
\right\|_1 .
\label{eq:loss_anchor}
\end{equation}
During training, the middle five frames of each Vimeo-90K septuplet form the medium-exposure reconstruction segment, while the remaining frames provide valid low-/high-exposure anchors. For $\mathcal{L}_{anc}$, $z_t^{(a)}$ and $z_t^{(b)}$ are obtained using two different valid anchor pairs randomly sampled from the same 7-frame training clip.
This auxiliary branch is used only during training. This encourages the fusion stage to preserve the structural content of the medium-exposure backbone and avoid over-dependence on a particular sparse anchor pair.

The overall objective is:
\begin{equation}
\mathcal{L}_{\mathrm{total}} = \mathcal{L}_{\mathrm{s}} + \lambda_{\mathrm{temp}}\,\mathcal{L}_{\mathrm{t}} + \lambda_{\mathrm{anc}}\,\mathcal{L}_{\mathrm{anc}}
\end{equation}
where $\lambda_{\mathrm{temp}}$ and $\lambda_{\mathrm{anc}}$ are the weights for the temporal difference loss and the anchor consistency loss, respectively.

\section{EXPERIMENTS}
\label{experiment}


\subsection{Experiment Settings}
\label{Experiment Settings}

\noindent \textbf{Training Details.}
We implement our model in PyTorch and train it on an NVIDIA RTX 4090 GPU. 
Optimization is performed using Adam~\cite{kingma2014adam} with an
initial learning rate of $10^{-4}$, which is decayed until it reaches
$10^{-6}$. 
During training, both LDR inputs and their corresponding HDR targets are
randomly cropped into $256\times256$ patches, with a batch size of 4.
Random rotations are applied for data augmentation to mitigate
overfitting.

\noindent \textbf{Datasets.}
Following previous HDR video reconstruction works \cite{xu2024hdrflow,chung2023lan,chen2021hdr}, we construct synthetic training data from Vimeo-90K septuplets \cite{xue2019video}. 
For each frame, we generate three LDR observations, denoted as $y_{1,t}$, $y_{2,t}$, and $y_{3,t}$, corresponding to low, medium, and high exposures. 
From the same observations, we construct two input protocols. 
Alternating-exposure baselines receive a single LDR stream sampled according to their original exposure patterns, while dual-stream setting uses a continuous medium-exposure stream $\mathbf{Y}_2=\{y_{2,t}\}_{t=1}^{T}$ together
with sparse low-/high-exposure anchors. 
Thus, all methods are evaluated from the same source clips, exposure settings, and ground-truth HDR frames, but under their respective capture assumptions.

For testing, we use 16 synthetic HDR sequences from the Cinematic Wide Gamut HDR-video and apply the same synthesis protocol to generate baseline and dual-stream inputs. 
We further evaluate on the real dual-camera dataset of \cite{zhang2025capturing}, which contains 22 scenes with over 200 frames each. 
Since real HDR ground truth is unavailable, we report no-reference metrics and qualitative results on this dataset.

\noindent \textbf{Evaluation metrics.}
We report $\textit{PSNR}_{\mu}$, $\textit{SSIM}_{\mu}$, and FovVideoVDP \cite{mantiuk2021fovvideovdp}, where $\textit{PSNR}_{\mu}$ and $\textit{SSIM}_{\mu}$ are computed in the $\mu$-law tone-mapped domain \cite{kalantari1}. 
For temporal evaluation, we use t-PSNR, t-SSIM, and the standard deviation (STD) of frame-wise reconstruction quality.

For a frame-wise metric $Q_t$, its temporal fluctuation is defined as
\begin{equation}
\mathrm{STD}(Q)
=
\sqrt{
\frac{1}{T}
\sum_{t=1}^{T}
(Q_t-\bar{Q})^2
},
\quad
\bar{Q}
=
\frac{1}{T}
\sum_{t=1}^{T}Q_t .
\end{equation}
Lower STD indicates a more stable temporal reconstruction.

\noindent \textbf{Baselines.}
Our method is evaluated against state-of-the-art video-based HDR reconstruction approaches, including HDRFlow~\cite{xu2024hdrflow}, LAN-HDR~\cite{chung2023lan}, and DeepHDRVideo~\cite{chen2021hdr}. 
To assess deghosting capability, we further include representative image-based HDR reconstruction methods, including AHDNet~\cite{yan2019attention}, HDRTransformer~\cite{liu2022ghost}, SCTNet~\cite{tel2023alignment}, DomainPlus~\cite{zheng2022domainplus}, AFUNet~\cite{li2025afunet} and EAFNet~\cite{zhang2025capturing}. 
For these image-based methods, we construct their multi-exposure inputs from the same dual-stream observations by using the current medium-exposure frame together with the corresponding sparse low-/high-exposure anchors within each reconstruction window. 
Methods based on few-shot or unsupervised learning are excluded, as they are not directly comparable under the supervised multi-exposure HDR video reconstruction setting.



\subsection{HDR Video Experimental Results}
\label{HDR Image Experimental Results}

\noindent \textbf{Quantitative Comparisons.}
Table~\ref{tab:main_syn} reports quantitative results on the synthetic HDR video benchmark. 
On slow/simple scenes, LoCAtion achieves the best temporal metrics and the lowest STD, while maintaining competitive spatial reconstruction quality. 
On dynamic/complex scenes, where large motion, saturation, and exposure-induced appearance changes are more prominent, LoCAtion consistently achieves the best PSNR$_\mu$, SSIM$_\mu$, t-PSNR, t-SSIM, and STD. 
These results indicate that the proposed backbone-guided compensation and sequence-level propagation are particularly beneficial under challenging dynamic conditions, where purely frame-centric reconstruction is more likely to suffer from correspondence errors and temporal flicker. Fig.~\ref{fig:speed_tradeoff} further compares the accuracy-efficiency trade-off. 
Although HDRFlow is faster, its reconstruction quality is lower, while Transformer-based baselines achieve competitive quality at substantially higher latency. 
LoCAtion provides a favorable balance by combining reliable cross-exposure routing with lightweight sequence propagation.

\noindent \textbf{Qualitative Comparisons.}
Fig.~\ref{figexp1} presents qualitative comparisons on representative challenging scenes. 
Image-based deghosting methods recover plausible local appearance but often leave residual ghosts around saturated or moving regions because each exposure stack is processed independently. 
Video-based HDR reconstruction methods reduce some ghosting artifacts, but their alignment modules may introduce noise or distorted structures under severe exposure changes and fast motion. 
In contrast, LoCAtion uses the continuous medium-exposure stream as a stable backbone and selectively routes reliable dynamic-range cues from sparse exposure anchors, producing cleaner highlights, sharper shadow details, and more stable local structures.
%

\begin{table}[t]
\centering
\caption{
Quantitative comparison with state-of-the-art methods on public real-captured HDR video datasets without ground-truth HDR references.
AE and DS denote alternating-exposure inputs and dual-stream inputs, respectively.
Rel. Time denotes the runtime ratio relative to our method.
Best and second-best results are highlighted in \best{bold} and \second{underline}, respectively.
}
 \label{tab:main_real}
\scriptsize
\setlength{\tabcolsep}{3.2pt}
\renewcommand{\arraystretch}{1.12}
\resizebox{0.5\textwidth}{!}{
\begin{tabular}{c c l ccc ccc}
\toprule
\multirow{2}{*}{\textbf{Input}}
&
\multirow{2}{*}{\textbf{Framework}}
&
\multirow{2}{*}{\textbf{Method}}
&
\multicolumn{3}{c}{\textbf{Real-Captured Quality}}
&
\multicolumn{3}{c}{\textbf{Efficiency}}
\\
\cmidrule(lr){4-6}
\cmidrule(lr){7-9}
&
&
&
\makecell[c]{LSD\\$\downarrow$}
&
\makecell[c]{$t$-SSIM\\$\uparrow$}
&
\makecell[c]{$t$-PSNR\\$\uparrow$}
&
\makecell[c]{Params\\(M) $\downarrow$}
&
\makecell[c]{Time\\(ms) $\downarrow$}
&
\makecell[c]{Rel. Time\\$\downarrow$}
\\
\midrule

\multirow{3}{*}{\makecell[c]{AE}}
&
\multirow{3}{*}{\makecell[c]{Frame-\\centric}}
&
DeepHDRVideo~(ICCV'21)~\cite{chen2021hdr}
&
14.0
&
0.7092
&
19.55
&
2.76
&
167.60
&
2.04$\times$
\\

&
&
LAN-HDR~(ICCV'23)~\cite{chung2023lan}
&
13.1
&
0.6527
&
20.26
&
7.39
&
230.70
&
2.81$\times$
\\

&
&
HDRFlow~(CVPR'24)~\cite{xu2024hdrflow}
&
11.9
&
0.7591
&
20.33
&
5.33
&
\best{17.60}
&
\best{0.21$\times$}
\\

\midrule

\multirow{7}{*}{\makecell[c]{DS}}
&
\multirow{6}{*}{\makecell[c]{Frame-\\centric}}
&
AHDNet~(CVPR'19)~\cite{yan2019attention}
&
\second{1.7}
&
0.8765
&
29.05
&
1.44
&
158.90
&
1.93$\times$
\\

&
&
HDR-Trans~(ECCV'22)~\cite{liu2022ghost}
&
\second{1.7}
&
0.8729
&
28.98
&
1.46
&
1531.30
&
18.64$\times$
\\

&
&
DomainPlus~(MM'22)~\cite{zheng2022domainplus}
&
2.5
&
0.8626
&
27.10
&
24.78
&
116.90
&
1.42$\times$
\\

&
&
SCTNet~(ICCV'23)~\cite{tel2023alignment}
&
1.8
&
0.8790
&
29.06
&
\best{0.96}
&
2553.90
&
31.08$\times$
\\

&
&
AFUNet~(ICCV'25)~\cite{li2025afunet}
&
\second{1.7}
&
0.8776
&
29.02
&
\second{1.14}
&
2341.50
&
28.50$\times$
\\

&
&
EAFNet~(arxiv'26)~\cite{zhang2025capturing}
&
\second{1.7}
&
\second{0.8794}
&
\second{29.07}
&
6.71
&
463.90
&
5.65$\times$
\\

\cmidrule(lr){2-9}

\rowcolor{gray!10}
&
\makecell[c]{Propagation}
&
\textbf{Ours}
&
\best{1.6}
&
\best{0.8806}
&
\best{29.12}
&
4.63
&
\second{82.16}
&
\second{1.00$\times$}
\\

\bottomrule
\end{tabular}
}
\vspace{1mm}
\end{table}

\begin{figure}[!t] 
\centering     \includegraphics[width=0.5\textwidth]{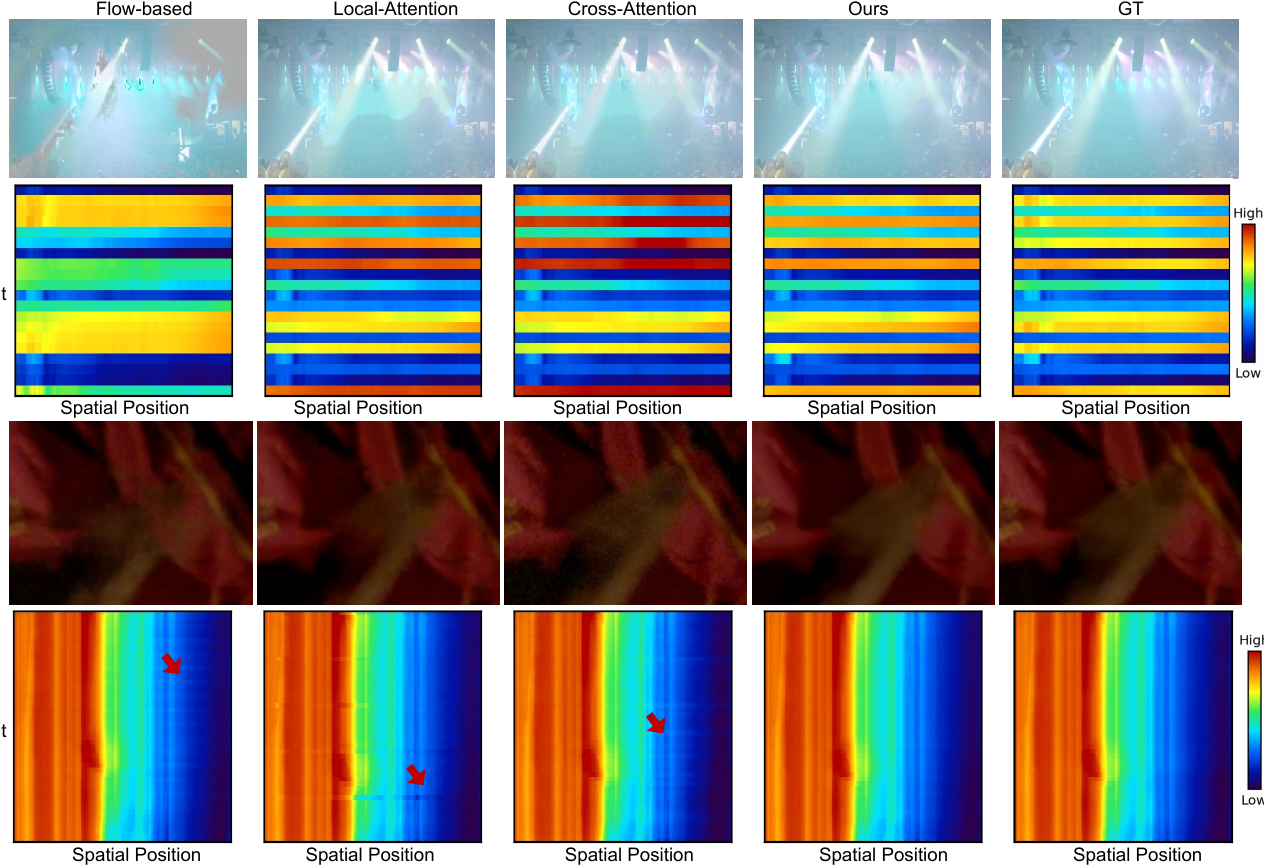} 
\caption{Comparison of cross-exposure routing strategies. 
All variants share the same dual-stream input and only replace the first-stage compensation module. 
Temporal profiles visualize scanline intensity evolution over time, where smoother profiles indicate better temporal consistency. 
Our CFA reduces artifacts and yields a more coherent temporal reconstruction.} 
\label{fig:alignment} 
 \end{figure}

\subsection{Real HDR Video Experimental Results}
\label{expvideo}

To assess real-world generalization, we evaluate LoCAtion on public real-captured HDR video sequences. 
Since ground-truth HDR frames are unavailable, we follow~\cite{zhang2025capturing} and report temporal PSNR, temporal SSIM, and luminance standard deviation (LSD). 
Here, t-PSNR and t-SSIM are computed between adjacent tone-mapped output frames to measure frame-to-frame intensity and structural consistency, while LSD measures temporal luminance fluctuation. 
Table~\ref{tab:main_real} reports the quantitative results, and the fourth row of Fig.~\ref{figexp1} provides representative visual comparisons on real-captured frames.

Alternating-exposure video methods suffer from unstable reference exposure and imperfect motion alignment on real sequences, often producing brightness flicker and residual ghosting around moving objects. 
Image-based deghosting methods use a fixed medium-exposure reference and therefore reduce global flicker, but they process each multi-exposure stack independently and lack temporal propagation, leaving local ghosting and detail fluctuations in high-contrast regions.

LoCAtion achieves the best temporal consistency on real videos, with the highest t-PSNR and t-SSIM and the lowest LSD. 
This improvement comes from two aspects: the continuous medium-exposure stream provides a stable structural backbone, and the proposed warping-free routing and sequence-level propagation suppress unreliable anchor cues while refining temporal details. 
As shown in the fourth row of Fig.~\ref{figexp1}, our method produces cleaner structures in real dynamic scenes. 
Additional visual comparisons and video demonstrations are available on the project page.


\begin{table}[t]
\centering
\caption{
Ablation studies of network components and design alternatives. 
All variants are trained and evaluated under the same dual-stream input protocol. 
The component-removal variants evaluate the contributions of DWT-domain fusion, TPA, BCA, and LRTM, while the design-alternative variants replace the proposed CFA stage with flow-based alignment, local attention, or cross-attention. 
}
\label{tab:component_ablation}
\scriptsize
\setlength{\tabcolsep}{2.6pt}
\renewcommand{\arraystretch}{1.10}
\resizebox{0.5\textwidth}{!}{
\begin{tabular}{lcccccc}
\toprule
\textbf{Variant}
&
\makecell[c]{STD\\$\downarrow$}
&
\makecell[c]{$PSNR_{\mu}$\\$\uparrow$}
&
\makecell[c]{$\Delta PSNR_{\mu}$\\$\downarrow$}
&
\makecell[c]{$SSIM_{\mu}$\\$\uparrow$}
&

\makecell[c]{Params\\(M) $\downarrow$}
&
\makecell[c]{Time\\(ms) $\downarrow$}
\\
\midrule

\multicolumn{7}{l}{\textit{Component removal}} \\
\midrule
w/o DWT
&
0.74
&
35.67
&
0.08
&
0.9048

&
4.50
&
220.8
\\

w/o ${\mathrm{TPA}}$
&
0.75
&
35.44
&
0.31
&
0.9004
&

3.78
&
61.3
\\

w/o ${\mathrm{BCA}}$
&
0.78
&
35.16
&
0.59
&
0.8969
&

2.64
&
80.2
\\

w/o ${\mathrm{LRTM}}$
&
0.73
&
35.51
&
0.24
&
0.9021
&

4.71
&
73.0
\\

w/o ${\mathrm{GSC}}$
&
0.82
&
34.62
&
1.13
&
0.8629
&

2.57
&
66.9

\\

\midrule
\multicolumn{7}{l}{\textit{Design alternatives}} \\
\midrule
Flow-based alignment
&
1.02
&
34.76
&
0.99
&
0.8774
&

4.17
&
153.2
\\

Local attention
&
0.80
&
35.39
&
0.36
&
0.9016
&

3.64
&
77.2
\\

Cross attention
&
0.74
&
35.53
&
0.22
&
\best{0.9051}
&

2.45
&
267.4
\\

\midrule
\rowcolor{gray!10}
\textbf{Ours}
&
\best{0.71}
&
\best{35.75}
&
--
&
\best{0.9051}
&
4.63
&
82.2
\\
\bottomrule
\end{tabular}
}
\vspace{-2mm}
\end{table}

\subsection{Ablation study}
\label{Ablation study}

We conduct ablation studies to validate the three key design choices of LoCAtion: 
(i) exploiting the temporal continuity of the medium-exposure sequence under sparse low-/high-exposure anchors, 
(ii) the design of Collaborative Feature Attention (CFA), including DWT-domain fusion and Temporal Prior Aggregator (TPA),
and (iii) the contribution of Global Sequence Consistency (GSC), including Bidirectional Context Aggregation (BCA) and Long-Range Temporal Modeling (LRTM).
Unless otherwise specified, all variants are trained and evaluated under the same dual-stream input protocol.

\begin{figure*}[t] 
\centering 
\includegraphics[width=1.0\textwidth]{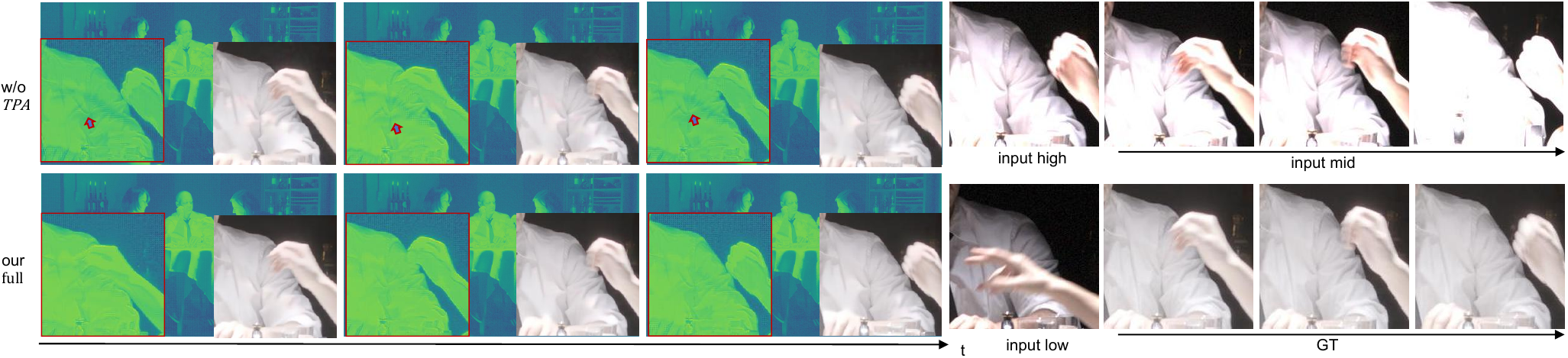} 
\caption{Ablation study on the impact of Temporal Prior Aggregator (TPA). Without TPA, the reconstructed details around moving regions tend to show duplication and ghosting-like artifacts due to unreliable exposure cue selection. With TPA, the motion region is reconstructed more consistently.} 
\label{fig:tpa_ablation} 
\end{figure*}

\noindent\textbf{Temporal input configuration.}
We first analyze whether sparse low-/high-exposure anchors are sufficient when coupled with a continuous medium-exposure backbone. 
Table~\ref{tab:temporal_input_ablation}(a) varies the medium-exposure sequence length $T$, which controls both the temporal context available to GSC and the temporal range over which sparse anchor information is propagated. 
When $T=1$, the model degenerates into frame-wise multi-exposure fusion, where each output is reconstructed from an isolated local stack without sequence-level propagation. 
This setting leads to inferior reconstruction quality and temporal stability, indicating that simply using dual-stream inputs without exploiting the continuous medium-exposure backbone is insufficient. 
Increasing the sequence length to $T=5$ clearly improves both PSNR$_{\mu}$ and STD, showing that a short continuous medium-exposure segment already provides effective temporal support for propagating sparse dynamic-range cues. 
Further increasing $T$ enlarges the effective temporal range and can improve some metrics, but it also requires propagating anchor information over longer distances and substantially increases memory consumption. 
For example, longer segments improve SSIM$_{\mu}$ in some cases, but the memory cost grows rapidly. 
We therefore adopt $T=5$ in the main experiments as a practical trade-off between reconstruction accuracy, temporal stability, and memory efficiency, rather than as an architectural limitation.

Table~\ref{tab:temporal_input_ablation}(b) further studies the sampling ratio between medium-exposure frames and low-/high-exposure anchor pairs. 
The 1:1 setting frequently changes the extreme-exposure anchors, which weakens temporal consistency and makes the reconstruction more sensitive to local anchor variations. 
In contrast, the 5:1 setting shares one anchor pair within a short medium-exposure backbone segment and achieves the lowest STD, indicating more stable sequence reconstruction. 
Sparser ratios such as 10:1 and beyond may improve frame-wise spatial metrics in some cases, but they noticeably increase temporal fluctuation because the dynamic-range guidance becomes too sparse and must be propagated over longer temporal distances. 
These results support our formulation: sparse exposure anchors are effective when propagated along a stable medium-exposure backbone, but both overly dense anchor switching and overly sparse anchor injection are suboptimal for temporally stable HDR reconstruction.

\begin{table}[t]
\centering
\caption{
Ablation study on temporal input configuration. 
(a) Effect of the medium-exposure sequence length $T$, which controls the temporal context available for propagation. 
(b) Effect of the medium-frame-to-anchor-pair sampling ratio. A ratio of 5:1 means that one low-/high-exposure anchor pair is shared by a 5-frame medium-exposure window. 
The selected configuration is highlighted in gray. 
}
\label{tab:temporal_input_ablation}
\scriptsize
\setlength{\tabcolsep}{3.6pt}
\renewcommand{\arraystretch}{1.0}
\resizebox{0.40\textwidth}{!}{
\begin{tabular}{lcccc}
\toprule
\multicolumn{5}{c}{\textbf{(a) Mid-Exposure Sequence Length}} \\
\midrule
\textbf{Setting}
&
\makecell[c]{STD\\$\downarrow$}
&
\makecell[c]{$PSNR_{\mu}$\\$\uparrow$}
&
\makecell[c]{$SSIM_{\mu}$\\$\uparrow$}
&
\makecell[c]{Memory\\(GB) $\downarrow$}
\\
\midrule

$T=1$
&
0.78
&
35.31
&
0.9012
&
4.79
\\

\rowcolor{gray!12}
$T=5$ \textbf{(Ours)}
&
0.71
&
\second{35.75}
&
0.9051
&
19.06
\\

$T=7$
&
\best{0.59}
&
\best{35.81}
&
0.9107
&
24.96
\\

$T=10$
&
0.62
&
35.73
&
0.9096
&
34.01
\\

$T=15$
&
\second{0.61}
&
35.66
&
\second{0.9115}
&
49.42
\\

$T=20$
&
0.64
&
35.57
&
\best{0.9125}
&
63.20
\\

\midrule[0.8pt]
\multicolumn{4}{c}{\textbf{(b) Number of Low/High Pairs}} \\
\midrule
\textbf{Setting}
&
\makecell[c]{STD\\$\downarrow$}
&
\makecell[c]{$PSNR_{\mu}$\\$\uparrow$}
&
\makecell[c]{$SSIM_{\mu}$\\$\uparrow$}
&
\makecell[c]{Memory\\(GB) $\downarrow$}
\\
\midrule

1-1
&
0.94
&
35.51
&
0.9024
& 19.06
\\

\rowcolor{gray!12}
5-1 \textbf{(Ours)}
&
\best{0.71}
&
35.75
&
0.9051
& 19.06
\\

10-1
&
1.23
&
\best{36.02}
&
\best{0.9115}
& 19.06
\\

15-1
&
1.18
&
\second{35.91}
&
\second{0.9100}
& 19.06
\\

20-1
&
1.16
&
35.71
&
0.9060
& 19.06
\\

55-1
&
\second{0.90}
&
35.47
&
0.9038
& 19.06
\\

\bottomrule
\end{tabular}
}
\end{table}

\noindent\textbf{Collaborative Feature Attention.}
To isolate the effect of cross-exposure routing, we replace CFA with representative alternatives, including flow-based alignment, local attention, and cross-attention, while keeping the remaining network unchanged. 
As shown in Table~\ref{tab:component_ablation} and Fig.~\ref{fig:alignment}, flow-based alignment suffers from exposure-induced correspondence errors, leading to the largest accuracy drop and unstable temporal profiles. 
Attention-based alternatives avoid explicit flow estimation, but still lack temporally informed reliability guidance and leave residual artifacts in saturated or fast-moving regions. 
By contrast, CFA achieves better reconstruction quality and more coherent temporal profiles by routing only reliable low-/high-exposure cues into the medium-exposure backbone.

\begin{figure*}[t] 
\centering 
\includegraphics[width=0.85\textwidth]{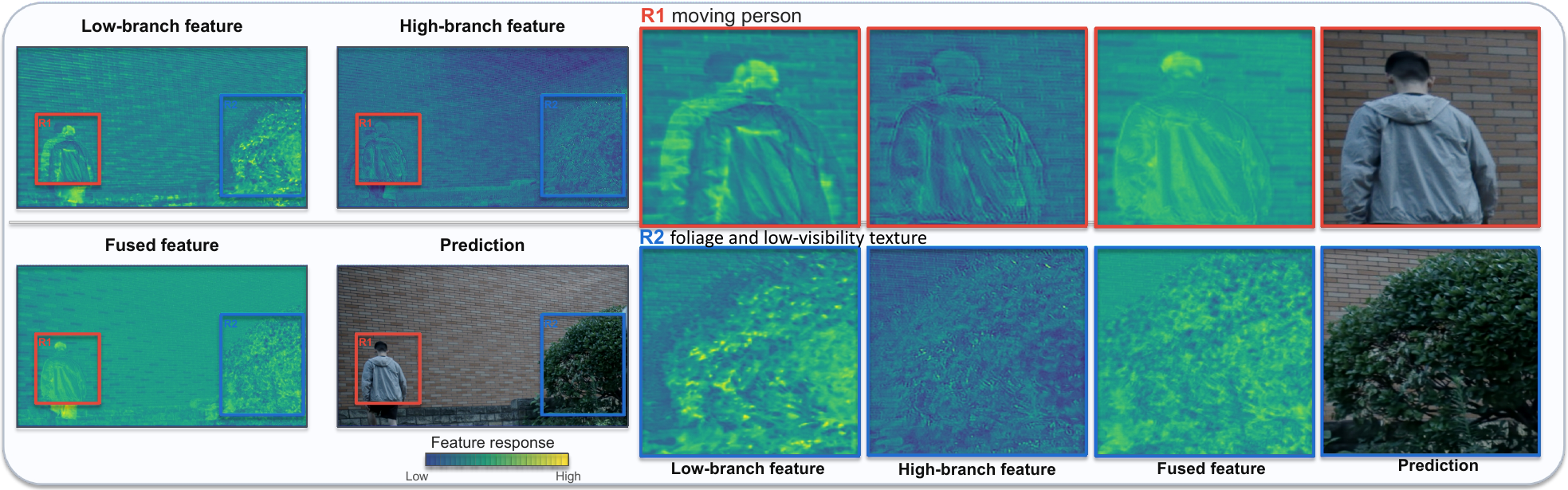} 
\caption{Feature-level analysis of CFA cross-exposure routing. 
We visualize normalized feature responses from the low-/high-exposure branches, their fused representation, and the final prediction. 
In $R1$, CFA suppresses inconsistent branch responses around motion, preventing duplicated structures. 
In $R2$, reliable complementary cues are retained to enhance subtle textures. 
Overall, CFA adaptively routes cross-exposure information for targeted fusion rather than naive feature merging.} 
\label{fig:cfa} 
\end{figure*}

The internal components of CFA are also important. 
Removing DWT-domain fusion substantially increases runtime, since cross-exposure aggregation is performed at full resolution. 
Removing TPA causes a clear degradation in both PSNR$_{\mu}$ and STD, demonstrating that temporally aggregated reliability priors are necessary for distinguishing useful dynamic-range compensation from motion-induced mismatch. 
The feature visualization in Fig.~\ref{fig:tpa_ablation} further confirms this behavior: without TPA, moving regions produce duplicated or smeared responses, whereas TPA yields more compact feature activations and suppresses unreliable anchor contributions. 
Fig.~\ref{fig:cfa} also shows that CFA does not naively merge exposure features; instead, it preserves reliable complementary details while suppressing inconsistent responses around motion boundaries.

\begin{figure*}[!t] 
\centering     \includegraphics[width=0.95\textwidth]{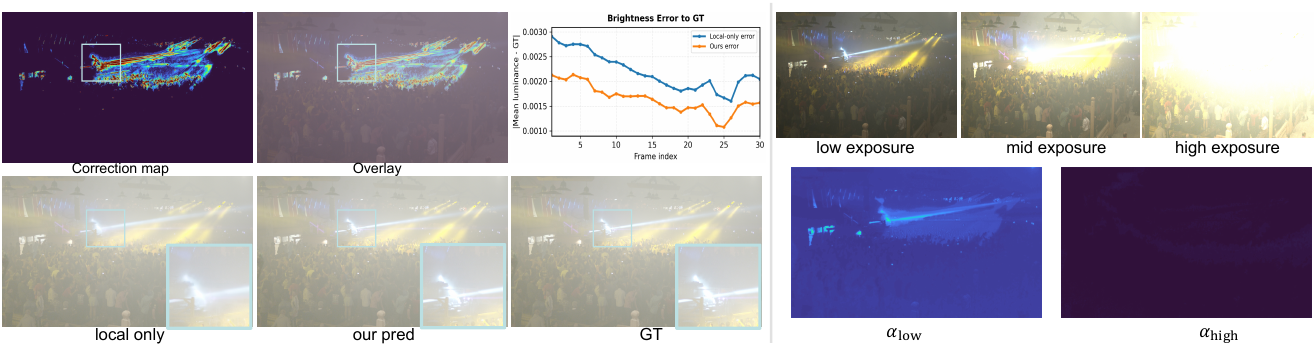} 
\caption{Reliability maps and GSC correction analysis. 
The low- and high-exposure anchors exhibit complementary reliability: \(\alpha_{\mathrm{low}}\) favors highlight-preserving features, while \(\alpha_{\mathrm{high}}\) is less trusted in saturated regions. 
Stage-1 (local-only) results are refined by GSC, where the residual concentrates on motion boundaries and other locally unstable areas. 
Compared with uniform smoothing, this targeted correction yields lower and temporally smoother brightness errors over the sequence.} 
\label{fig:feature_map} 
 \end{figure*}

\noindent\textbf{Global Sequence Consistency.}
As reported in Table~\ref{tab:component_ablation}, removing BCA results in a notable drop in reconstruction quality and temporal stability, indicating that bidirectional propagation is essential for refining local Stage-1 estimates with both past and future context. 
Removing LRTM also degrades performance, showing that long-range dependency modeling complements local bidirectional aggregation by capturing sequence-level consistency beyond the immediate temporal window.

Fig.~\ref{fig:feature_map} provides a more detailed visualization of the refinement behavior. 
The GSC residual is mainly concentrated around motion boundaries, saturated highlights, and locally unstable regions, while already stable areas receive much weaker correction. 
This indicates that GSC does not simply smooth the entire sequence. 
Instead, it performs content-adaptive temporal correction guided by bidirectional and long-range context. 
The corresponding brightness-error curve further shows lower and smoother errors after GSC, confirming that the proposed temporal refinement improves both reconstruction fidelity and frame-to-frame stability.

\section{Conclusion}
We presented LoCAtion, a continuous-backbone-driven framework for dual-stream HDR video reconstruction. By converting the hardware-level continuity of the medium-exposure stream into an algorithmic prior, LoCAtion organizes HDR recovery as backbone-guided cross-exposure compensation rather than independent reference-centered fusion. The medium-exposure stream maintains temporally coherent scene structure, while sparse low-/high-exposure anchors provide complementary dynamic-range cues. Collaborative Feature Attention selectively routes reliable cross-exposure cues into the backbone without explicit optical-flow-based warping, and Global Sequence Consistency refines local HDR estimates through bidirectional sequence-level propagation. Experiments on synthetic benchmarks and real-captured videos show that LoCAtion improves reconstruction quality and temporal stability, particularly in fast-motion and complex scenes, while maintaining a favorable accuracy-efficiency trade-off.

\bibliographystyle{IEEEtran}
\bibliography{ref}

\end{document}